\documentclass[runningheads]{llncs}
\usepackage[T1]{fontenc}
\usepackage{graphicx}
\usepackage{xcolor}
\usepackage{amsmath}
\usepackage{hyperref}

\usepackage{amssymb}
\usepackage{enumitem}
\usepackage{float}
\usepackage{physics}
\usepackage{booktabs}
\usepackage{listings}

\lstdefinestyle{Python}{
    language        = Python,
    frame           = lines, 
    basicstyle      = \footnotesize,
    keywordstyle    = \color{blue},
    stringstyle     = \color{green},
    commentstyle    = \color{red}\ttfamily
}

\begin{document}

\title{CBAGAN-RRT: Convolutional Block Attention Generative Adversarial Network for Sampling-Based Path Planning}

\author{
Abhinav Sagar\inst{1} \and
Sai Teja Gilukara\inst{2}
}

\institute{
University of Maryland, College Park, Maryland, USA\\
\email{asagar@umd.edu}
\and
University of Maryland, College Park, Maryland, USA\\
\email{saitejag@umd.edu}
}

\maketitle              

\begin{abstract}

Sampling-based path planning algorithms play an important role in autonomous robotics. However, a common problem among these algorithms is that the initial path generated is not optimal, and the convergence is too slow for real-world applications. In this paper, we propose a novel image-based learning algorithm using a Convolutional Block Attention Generative Adversarial Network (CBAGAN-RRT) with a combination of spatial and channel attention and a novel loss function to design the heuristics, find a better optimal path, and improve the convergence of the algorithm, both concerning time and speed. The probability distribution of the paths generated from our GAN model is used to guide the sampling process for the RRT algorithm. We demonstrate that our algorithm outperforms the previous state-of-the-art algorithms using both the image quality generation metrics, like IOU Score, Dice Score, FID score, and path planning metrics like time cost and the number of nodes. Ablation studies show the effectiveness of various components in our network architecture. The advantage of our approach is that we can avoid the complicated preprocessing in the state space, our model can be generalized to complex environments like those containing turns and narrow passages without loss of accuracy, and our model can be easily integrated with other sampling-based path planning algorithms.

\end{abstract}

\section{Introduction}

Path planning is one of the fundamental topics in robotics that aims to find a collision-free path around environmental obstacles. Various approaches have been proposed over the years, like grid-based ones using Dijkstra's and A*, which suffer from the high time and memory cost as the dimensionality of the search space increases. Sampling-based path planning algorithm solves the aforementioned problem by drawing samples from the state space, constructing a random graph, and searching for a feasible path using the Probabilistic Roadmap (PRM) or the Rapidly Exploring Random Tree (RRT), thus offering better scalability to higher-dimensional problems. A random graph is constructed in PRM after the sampling process and hence works better for multi-query problems, while the random tree is constructed randomly and incrementally from the start to the goal node, thus making it better for single-query problems. Moreover, RRT and the PRM-based algorithms guarantee that a solution is found if it exists because of the probabilistic completeness property. 

However, sampling-based path planning algorithms' convergence speed is slow, and the initial paths generated are not optimal because they sample uniformly in the state space and have many time-consuming and repeated functions like collision checking and the tree rewiring operation in the case of the RRT* algorithm. The path quality generated is proven to be not optimal, and these algorithms are too slow in challenging environments like those having many turns or narrow passages. In this work, we propose a novel GAN-based neural network that can quickly predict the probability distribution of feasible paths on a map before being fed to the sampling-based path planning algorithm, thus speeding up the convergence both in terms of time and speed.

\section{Related Work}

Sampling-based path planning algorithms \cite{lavalle2006planning} have been highly successful in various robotics problems, from autonomous driving to drones and warehouse robots. The first sampling-based path planning algorithm was the original RRT algorithm by \cite{lavalle1998rapidly}, which solved several challenges faced by the grid-based algorithms, like faster convergence and a better initial solution. It was considered the perfect sweet spot between probabilistic road maps and grid-based path planning algorithms, getting the best of both worlds as shown in \cite{lavalle2004relationship} and \cite{lavalle2001randomized}. 

\cite{karaman2011sampling} showed that the RRT* algorithm possesses an asymptotically optimal property, and the quality of the path generated was much better than the RRT algorithm using a novel rewiring operation and a novel nearest neighbor search operation. However, the algorithm was much slower than the RRT algorithm, and it took a long time to converge, and the paths obtained were not always correct. The RRT* algorithm works under holonomic constraints, offline planning mode, point kinematic model, uniform sampling strategy, and Euclidean term as the distance metric. 

\cite{karaman2011anytime} proposed Anytime RRT*, which was able to generate better trajectories using a branch and bound adaptation technique and a committed trajectory technique, and was able to generate better optimal paths while also speeding up the convergence, thus demanding fewer computational resources. However, the paths generated were still not optimal in a lot of cases, and the algorithm forced node removal while building the tree.  Anytime RRT* algorithm works under non-holonomic constraints, online planning mode, Dublin car kinematic model, uniform sampling strategy, and a Euclidean plus a velocity term as the distance metric. 

\cite{nasir2013rrt} proposed RRT* Smart, which improved the efficiency and accelerated the rate of the convergence of the algorithm using a novel path optimization technique and intelligent sampling. However, this method had its drawbacks, like manually defining the heuristics upon which the algorithm depended, and there was a trade-off between exploration space and the convergence rate. The RRT* Smart algorithm works under holonomic constraints, offline planning mode, point kinematic model, intelligent sampling strategy, and Euclidean term as the distance metric. 

Machine learning-based approaches have been applied and have had their share of success by either improving the heuristics to speed up the convergence of the algorithm or avoiding getting stuck in the local minimum. Deep learning powered by deep neural networks has ushered a new era in this domain and has been quite successful in solving various problems in sampling-based path planning \cite{wang2021deep}, \cite{wang2022gmr}, \cite{mcmahon2022survey}, \cite{arslan2015machine}, \cite{burns2005sampling}, \cite{yu2021reducing}, \cite{li2018neural}, \cite{dang2022deep}. 

Recent years have also seen an advent of generative models like the Generative Adversarial Network like Pix2pix \cite{isola2017image} and Variational Autoencoder to generate the promising region using paired labeled training data and have shown to outperform all previous numerical and algorithmic-based approaches while reducing the time taken, number of nodes, and the length of the initial path to convergence \cite{zhang2021generative}, \cite{ma2022enhance}, and \cite{ma2021conditional}. Attention mechanism \cite{vaswani2017attention} has resulted in a breakthrough and has been a driving force behind various state-of-the-art architectures across modalities like computer vision and natural language processing \cite{zhang2019self}, \cite{wang2020eca}, \cite{woo2018cbam}, \cite{fu2019dual}, and \cite{fu2019dual} and offers great potential in the image based path planning landscape.

\section{Method}

\subsection{Path Planning Problem}

Assuming the state space of the planning problem is represented by X, and $X_{free}$ is the collision-free subspace of the state X. Then $X_obs = X - X_{free}$ would denote the obstacle space in the map. Let's assume the start and the goal state are represented by $x_s$ and $x_g$, and any 2 random points by $x_1$ and $x_2$ in the map. Let's also denote the ball center of radius r at x by $B(x, r)$. The Euclidean distance between the 2 can be denoted by $x_1 x_2$. Assuming $\sigma$ denotes a feasible path and for the sequence of states we have $\sigma(0)$ = $x_s$, $\sigma(t)$ = $B(x_g, r_g$ where $\sigma(t) \subset X_{free}$. Hence, the cost of the optimal path can be represented by $c (\sigma) = \Sigma_{t=1}^t \sigma(t) - \sigma(t-1)$. The path length can be obtained by minimizing the cost function. The following definition is used in the RRT algorithm:

- V: Set of vertices in the sampling tree.
- E: Set of edges between the vertices in the sampling tree.
- Sample free: Procedure of sampling a random node from the state space.
- Nearest(V, x): Finding the nearest node from the vertex V to point X using Euclidean distance as the metric.
- Steer($x_1$, $x_2$): Steer from $x_1$ to $x_2$ along the path $\delta (t)$.
- $Obs_{free}\delta(t)$: Find if a collision-free and feasible path exists in the map or not.

\subsection{Dataset}

We used the dataset generated by \cite{zhang2021generative} to validate our results. The dataset was generated by randomly placing different obstacles on the map and randomly sampling the start and goal nodes, which are denoted by red and blue dots on the map, respectively. The RRT algorithm was run to generate the feasible path, which is shown in green color, or the ground truth. The dimensions of all the images are ($3\times256\times256$), where the height and the width of the images are 256, and the number of channels is 3. We use 8000 images for training and 2000 images for testing, respectively. There are 200 different types of environment maps present in the dataset.

\subsection{Data Augmentation}

We use the following data augmentation operations on the map as shown in \autoref{data_augmentation}:

\begin{table}[h]
  \caption{Data Augmentation Parameters}
  \label{sample-table1}
  \centering
  \begin{tabular}{ll}
  \toprule
    Parameter     & Value           \\
    \midrule
    Height shift of map            & 2  \\
    Width shift of map   & 2         \\
    Shift step of map          & 1       \\
     Rotation probability of map & 0.5\\
     Number of maps to be generated & 10\\
    \bottomrule
  \end{tabular}
  \label{data_augmentation}
\end{table}

The following data augmentation methods were used to increase the quality and quantity of the dataset:

\begin{itemize}

\item{Rescaling:} We rescale the pixel values by
a rescaling factor of 1/255. 

\item{Rotation:} Random rotations with the setup degree range between [0, 360] were used.

\item{Height and Width Shift:} Shifting the input to the left or right and up or down was performed.

\item{Shearing Intensity: It refers to the shear angle (unit in degrees) in a counter-clockwise direction.}

\item{Brightness:} It uses a brightness shift value from the setup range.

\end{itemize}

\begin{figure}[h]
    \centering
    \includegraphics[width=6cm]{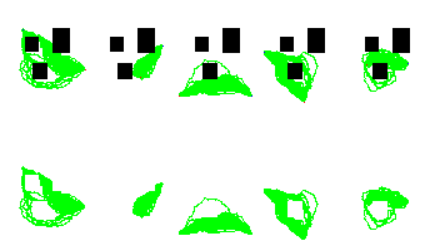}
    \caption{Illustration of ground truth images with promising regions in the environment map. The first row shows images from the dataset with the white region as the free space, the black region as the obstacle space, the start node in color red, the goal node in color blue, and the green area as the promising region superimposed on the map. The second row shows only the promising region on the map. The promising region or the region of interest here is the ground truth labels.}
    \label{sample_dataset}
\end{figure}

A few sample ground truth images from the dataset generated by \cite{zhang2021generative}, showing the promising region in the environment map as shown in \autoref{sample_dataset}

\subsection{Network Architecture}

The fundamental goal in this image-based heuristic generation for the environment is to map an RGB image $X \in R^{H\times W\times 3}$ to a semantic map $Y \in R^{H\times W\times C}$ with the same spatial resolution $H \times W$, where $C$ is the number of classes of objects present in image which is 2, in this case, comprised of free space and the obstacle space. The input image $X$ is converted to a set of feature maps ${F_{l}}$ where l=1,...,3 from each network stage, where $F_{l} \in R^{H_{l}×W_{l}×C_{l}}$ is a $C_{l}$-dimensional feature map. The input image is first passed through a block comprising a convolutional layer, a batch normalization layer, and a ReLU activation function. We express a convolution layer $W^{n}(x)$ depicted as in \autoref{convolutional_layer}:

\begin{equation}
\mathrm{W}^{n}(x)=\mathbf{W}^{n \times n} \odot x+\mathbf{b}
\label{convolutional_layer}
\end{equation}

where $\odot$ represents the convolution operator, $W^{n\times n}$ represents the $n \times n$ convolutional kernel, $x$ represents the input data, and $b$ represents the bias vector.

\subsubsection{Spatial Attention Module}

The spatial attention module is used for capturing the spatial dependencies of the feature maps. The spatial attention (SA) module in our network is defined as in \autoref{spatial_attention_eqn}:

\begin{equation}
f_{{SA}}(x)=f_{sigmoid}\left({W}_{2}\left(f_{{ReLU}}\left({W}_{1}(x)\right)\right)\right)
\label{spatial_attention_eqn}
\end{equation}

where $W_{1}$ and $W_{2}$ denotes the first and second $1 \times 1$ convolution layer respectively, $x$ denotes the input data, $f_{Sigmoid}$ denotes the sigmoid function, $f_{ReLU}$ denotes the ReLU activation function. The spatial attention module used in this work is shown in \autoref{spatial_attention}:

\begin{figure}[htp]
    \centering
    \includegraphics[width=7cm]{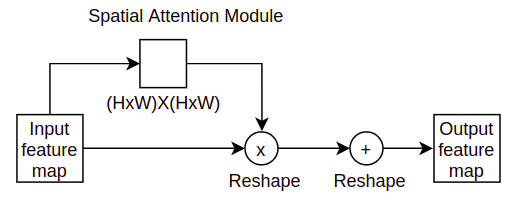}
    \caption{Illustration of the spatial attention module}.
    \label{spatial_attention}
\end{figure}

\subsubsection{Channel Attention Module}

The channel attention module is used for extracting high-level semantic information. The channel attention (CA) module in our network is defined as in \autoref{channel_attention_eqn}:

\begin{equation}
f_{{CA}}(x)=f_{sigmoid }({W}_{2}(f_{{ReLU }}({W}_{1}f_{{AvgPool }}^{1}(x))))
\label{channel_attention_eqn}
\end{equation}

where $W_{1}$ and $W_{2}$ denotes the first and second $1 \times 1$ convolution layer, $x$ denotes the input data. $f^{1}_{AvgPool}$ denotes the global average pooling function, $f_{Sigmoid}$ denotes the Sigmoid function, $f_{ReLU}$ denotes ReLU activation function. The channel attention module used in this work is shown in \autoref{channel_attention}:

\begin{figure}[htp]
    \centering
    \includegraphics[width=7cm]{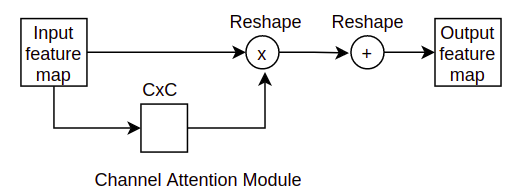}
    \caption{Illustration of the channel attention module}.
    \label{channel_attention}
\end{figure}

\subsubsection{Attention Block}

Assuming an input feature map $F$, the final (channel and spatial) attention-enhanced feature map is computed by element-wise multiplying $F$ with both the channel and spatial attention maps, as shown in~\autoref{ca} and~\autoref{sa}~\cite{woo2018cbam}: 

\begin{equation}
F_c=f_{{CA}}\left(F\right) \otimes F, 
\label{ca}
\end{equation}

\begin{equation}
F_o=f_{{SA}}\left(F_c\right) \otimes F_c.
\label{sa}
\end{equation}

Here $\otimes$ denotes element-wise multiplication and $F_o$ is the final attention-enhanced feature map. 

The attention block, comprising channel attention and spatial attention, is shown in \autoref{attention_block}:

\begin{figure}[htp]
    \centering
    \includegraphics[width=8cm]{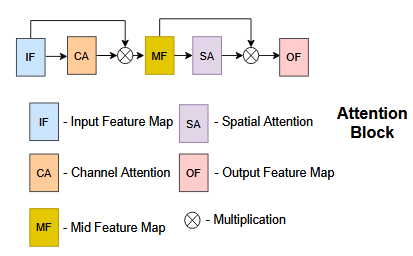}
    \caption{Illustration of the attention block in the point and map discriminator.}.
    \label{attention_block}
\end{figure}

\subsubsection{Convolutional Block Attention Generative Adversarial Network}

GANs are a family of unsupervised generative models that learns to generate samples from a given distribution, originally proposed in \cite{goodfellow2014generative}. Given a noise distribution, Generator $G$ tries to generate samples while the Discriminator $D$ tries to tell whether the generated samples are from the correct distribution or not. Both the generator and discriminator are trying to fool each other, thus playing a zero-sum game. In other words, both are in a state of Nash Equilibrium. Let $G$ represent the generator, $D$ the discriminator, the loss function used for training GAN is depicted as in \autoref{loss_gan}:

\begin{multline}
\mathcal{F}(\mathcal{D}, \mathcal{G})=\mathbb{E}_{\mathbf{x} \sim p_{\mathbf{x}}}[-\log \mathcal{D}(\mathbf{x})]+ \mathbb{E}_{\mathbf{z} \sim p_{\mathbf{z}}}[-\log (1-\mathcal{D}(\mathcal{G}(\mathbf{z})))]
\label{loss_gan}
\end{multline}

where $z$ is latent vector, $x$ is data sample, $p_{z}$ is probability distribution over latent space and $p_{x}$ is the probability distribution over data samples. The zero-sum condition is defined in \autoref{zero_sum}:

\begin{equation}
\min _{G} \max _{D} \mathcal{F}(\mathcal{D}, \mathcal{G})
\label{zero_sum}
\end{equation}

In our case, the generator takes points, map, and noise vectors as input, and all three vectors are passed through a convolutional layer with input channels of 3, 3, and 1 and output channels of 16, 16, and 32, respectively, for points, map, and noise vectors. Each of these convolutional layers has a kernel size of 3, a stride of 1, and ReLU is used as an activation function. The 3 intermediate feature maps obtained are concatenated and fed to four residual blocks for downsampling first, and then followed by four residual blocks for upsampling. The last layer is a convolutional layer to output three channels with a kernel size of 1, a stride of 1, zero padding, and tanh as an activation function.

Two discriminators are used in this work: the map discriminator and the point discriminator. The only difference between the two discriminators is that the map discriminator takes the map and the Region of Interest (ROI) vectors as input, while the point discriminator takes the point and the ROI vectors as input. The map/point and the ROI vectors are both initially passed through a convolutional layer with input and output channels of 3 and 32, respectively, and leaky ReLU as an activation function, followed by the attention block (A residual connection is used to stabilize training by connecting the convolutional layer output to the attention block output). The two outputs are concatenated and fed to two convolutional layers (the channel size increased by two at every layer, Leaky ReLU is used as an activation function along with batch normalization), followed by the residual connection-based attention block, and finally followed by two more convolutional layers, the last one resulting in one channel as output with a kernel size of two.

The generator and discriminator network architectural details are shown in \autoref{generator_discriminator}:

\begin{figure}[htp]
    \centering
    \includegraphics[width=10cm]{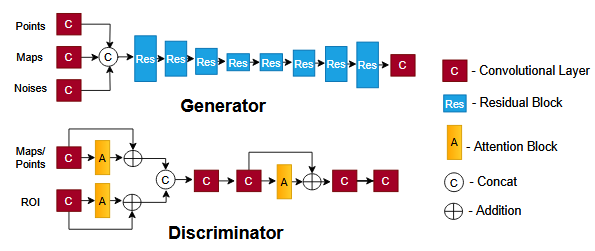}
    \caption{Illustration of the generator and discriminator.}.
    \label{generator_discriminator}
\end{figure}

\begin{figure}[htp]
    \centering
    \includegraphics[width=10cm]{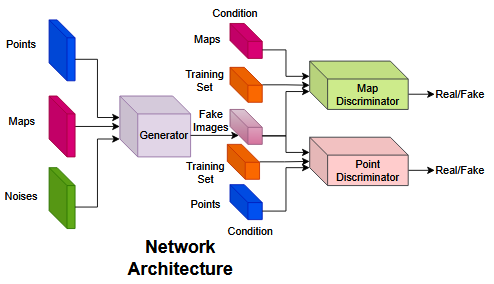}
    \caption{Illustration of the complete network architecture.}.
    \label{network_architecture}
\end{figure}

The generator takes points, maps, and noise vectors as input, and produces fake images or the ROI in this case as output. The map discriminator takes the map vector and the ROI vector as input, while the point discriminator takes the point vector and the ROI vector as input and outputs whether the generated image is real or fake. The complete network architecture used in our work is shown in \autoref{network_architecture}.

\subsection{Loss Functions}

A combination of binary cross-entropy and dice loss has been used to train the network. The first part of binary cross-entropy is a commonly used loss function for classification problems, as shown by \cite{goodfellow2016deep}. Every pixel in the image needs to be classified, and hence the loss function can be written as shown in \autoref{ce_loss}:

\begin{equation}
\mathcal{L}_{C E}=-\sum_{i, j} y_{i, j} \log \widehat{y}_{i, j}+\left(1-y_{i, j}\right) \log \left(1-\widehat{y}_{i, j}\right)
\label{ce_loss}
\end{equation}

The first of them is a binary cross-entropy loss in the x-direction as shown in \autoref{bce_x}:

\begin{multline}
L_{b c e}^x\left(\mathcal{P}^x, \hat{\mathcal{P}}^x\right)=\sum_{i=1}^H \sum_{j=1}^W \mathcal{P}_{i, j}^x \log \left(\hat{\mathcal{P}}_{i, j}^x\right)+ \left(1-\mathcal{P}_{i, j}^x\right) \log \left(1-\hat{\mathcal{P}}_{i, j}^x\right)
\label{bce_x}
\end{multline}

A similar binary cross-entropy loss term is also used in the y-direction as shown in \autoref{bce_y}:

\begin{multline}
L_{b c e}^y\left(\mathcal{P}^y, \hat{\mathcal{P}}^y\right)=\sum_{i=1}^H \sum_{j=1}^W \mathcal{P}_{i, j}^y \log \left(\hat{\mathcal{P}}_{i, j}^y\right)+ \left(1-\mathcal{P}_{i, j}^y\right) \log \left(1-\hat{\mathcal{P}}_{i, j}^y\right)
\label{bce_y}
\end{multline}

The total binary cross-entropy loss term can be obtained by summing up the two individual terms as shown in \autoref{total_ce_loss}:

\begin{equation}
L_{b c e}^{x y}(\mathcal{P}, \hat{\mathcal{P}})=L_{b c e}^x\left(\mathcal{P}^x, \hat{\mathcal{P}}^x\right)+L_{b c e}^y\left(\mathcal{P}^y, \hat{\mathcal{P}}^y\right)
\label{total_ce_loss}
\end{equation}

The problem with binary cross-entropy loss is that it doesn't take into account the class imbalance, as the background is typically the dominant class in this case. This is one of the fundamental challenges in image segmentation problems. Dice Loss is robust to the aforementioned problem, which is based on Dice Similarity Coefficient as defined in \autoref{dice_loss}:

\begin{equation}
L_{\text {dice }}^{x y}(\mathcal{P}, \hat{\mathcal{P}})=1-\frac{2\left|\mathcal{P}^x \cap \hat{\mathcal{P}}^x\right|+2\left|\mathcal{P}^y \cap \hat{\mathcal{P}}^y\right|}{\left|\mathcal{P}^{x 2}\right|+\left|\hat{\mathcal{P}}^{x 2}\right|+\left|\mathcal{P}^{y 2}\right|+\left|\hat{\mathcal{P}}^{y 2}\right|}
\label{dice_loss}
\end{equation}

Hence, the total loss can be obtained, which is a summation of the binary cross-entropy loss and dice loss, which is used for training the model, and is shown in \autoref{total_loss}:

\begin{equation}
L(\mathcal{P}, \hat{\mathcal{P}})=L_{\text {bce }}^{x y}(\mathcal{P}, \hat{\mathcal{P}})+L_{\text {dice }}^{x y}(\mathcal{P}, \hat{\mathcal{P}})
\label{total_loss}
\end{equation}

The loss function used for training the map discriminator can be represented as shown in \autoref{loss_map_discriminator}:

\begin{equation}
\begin{aligned}
\mathcal{L}_{D_{\text {map }}}= L(\mathcal{P}, \hat{\mathcal{P}}) + & \mathbb{E}\left[\log D_{\text {map }}(s, m)\right]+ \\
& \left.\mathbb{E}_{[} \log \left(1-D_{\text {map }}(G(z, m, p), m)\right)\right]
\end{aligned}
\label{loss_map_discriminator}
\end{equation}

The loss function used for training the point discriminator can be represented as shown in \autoref{loss_point_discriminator}:

\begin{equation}
\begin{aligned}
\mathcal{L}_{D_{\text {point }}}= L(\mathcal{P}, \hat{\mathcal{P}}) + & \mathbb{E}\left[\log D_{\text {point }}(s, p)\right]+ \\
& \left.\mathbb{E}_{[} \log \left(1-D_{\text {point }}(G(z, m, p), p)\right)\right] 
\end{aligned}
\label{loss_point_discriminator}
\end{equation}

The mean square pixel-wise loss function is defined in \autoref{mse_loss}:

\begin{equation}
\mathcal{L}_{mse}(X, Y)=\ell_{mse}(G(X), Y)=\|G(X)-Y\|^{2}
\label{mse_loss}
\end{equation}

Hence, the loss function used for training the generator can be represented as in \autoref{total_generator_loss}:

\begin{equation}
\begin{aligned}
\mathcal{L}_G = & \alpha_1 \mathbb{E}\left[\log D_{\text {map }}(G(z, m, p), m)\right]+ \\ & \alpha_2 \mathbb{E}\left[\log D_{\text {point }}(G(z, m, p), p)\right] + \mathcal{L}_{mse}(X, Y) 
\end{aligned}
\label{total_generator_loss}
\end{equation}

where $\alpha_1$ and $\alpha_2$ are both 0.5.

\subsection{Path Planning Algorithm}

The RRT algorithm is used for single-query applications and is suitable for dealing with high-dimensional problems. The algorithm incrementally builds a tree of feasible paths rooted at a start node by randomly expanding nodes in a collision-free space. 

Initially, there is an empty vertex set V, an edge set E, and an initial state. At every iteration, a node is randomly sampled and the nearest node of the existing vertex sex V is connected to the new sampling node. The vertex set and the edge set are added if the connection works. The process keeps on repeating until a new sampling node in the goal region is contained in the expanding tree. Space-filling trees are constructed to search a path connecting start and goal nodes while incrementally drawing random samples from the free space. 

We generate the promising regions using GAN to create the heuristics, which denotes that a feasible path exists with a high probability. RRT is utilized as a sampling-based path-planning algorithm. 

We use our pre-trained GAN model to guide the sampling process of the RRT algorithm. The algorithm attempts to find the nearest vertex to the new sample and connect them, and the new vertex is added to the vertex set V, and the edge is added to the edge set E if no obstacles are present in the connection. Finally, the algorithm returns a new vertex set V and edge set E. 

The Heuristic RRT algorithm used in this work is shown below:\\

    \lstset{caption={RRT Algorithm}}
    \lstset{label={lst:alg1}}
     \begin{lstlisting}[style = Python]
RRT(Map, start, goal)
  qs = [start]
  qs_parent = [0]
  while True{
    q_rand = random_vertex_generated()
    n_idx,q_near = nearest_vertex_check(q_rand)
    q_new = new_point_generate(q_near,
                                q_rand,
                                n_idx)
    if connection_check(q_new){
      break
    }
  }
    \end{lstlisting}
    
\subsection{Training Details}

We train the model on an Nvidia T4 GPU in Pytorch using mini-batch Stochastic Gradient Descent (Adam) as the optimizer. The momentum and weight decay of the optimizer are set to 0.9 and 0.0001, respectively. We use exponential decay as the learning rate scheduler. We train the model for 50 epochs 8 with a batch size. The input maps and the ground truth feasible region map are combined and sent to training. The exponential learning rate scheduler and ReduceLROnPlateau were used. The hyperparameter values for the Adam optimizer are ($\beta_{1}$ = 0.5, $\beta_{2}$ = 0.999.

\subsection{Evaluation Metrics}

We use the time cost, which represents the time taken for the algorithm to find a solution, and the number of nodes, which denotes the number of nodes explored on the way to the solution, as the path planning metrics to evaluate the quality of our algorithm from the path planning perspective. Dice Score (Dice), also known as F1-score, and Intersection over Union (IoU) are utilized as the image quality metrics to evaluate the performance of our network. Dice Score and IOU scores are computed as shown in \autoref{dice_metric} and \autoref{iou_metric}:

\begin{equation}
\mathrm{Dice}=\frac{2 T P}{2 T P+F N+F P}
\label{dice_metric}
\end{equation}

\begin{equation}
\mathrm{IoU}=\frac{T P}{T P+F N+F P}
\label{iou_metric}
\end{equation}

Here, True positive (TP), false negative (FN), and false positive (FP) numbers of pixels can be calculated separately for each image and averaged over the test set. 

We also use Fréchet Inception Distance (FID) to evaluate the quality of generated promising regions. A lower FID is preferred for better-performing generative models. Let $D$ represent the CNN used to extract features, ($m_{r}$, $\sigma_{r}$) be the mean and covariance of features extracted from real samples, and ($m_{f}$, $\sigma_{f}$) be mean and covariance of features extracted from fake samples with $D$, then the Frechet Inception distance is defined using \autoref{fid_score}:

\begin{multline}
d^{2}\left(\left(m_{r}, \Sigma_{r}\right),\left(m_{f}, \Sigma_{f}\right)\right)=\left\|m_{r}-m_{f}\right\|_{2}^{2}+\operatorname{Tr}\left(\Sigma_{r}+\Sigma_{f}-2\left(\Sigma_{r} \Sigma_{f}\right)^{1 / 2}\right)
\label{fid_score}
\end{multline}

\subsection{Experimental Details}

The hyperparameters used in our experiments are shown as in \autoref{hyperparameters}:

\begin{table}[h]

  \caption{Hyperparameters}
  \label{sample-table1}
  \centering
  \begin{tabular}{ll}
  \toprule
    Parameter     & Value           \\
    \midrule
    Batch Size             & 8           \\
    Epochs             & 20            \\
     Generator Learning Rate & 0.0001\\
Map Discriminator Learning Rate  & 0.00005\\
Point Discriminator Learning Rate& 0.00005 \\
    Optimizer & Adam \\
    \bottomrule
  \end{tabular}
  \label{hyperparameters}
\end{table}

The parameters associated with the various loss functions are shown as in \autoref{loss_function_parameters}:

\begin{table}[h]

  \caption{Loss Function Parameters}
  \label{sample-table1}
  \centering
  \begin{tabular}{ll}
  \toprule
    Parameter     & Value           \\
    \midrule
    $\alpha$ in Dice loss            & 10           \\
    The smooth parameter in Dice loss  & 1         \\
     $\alpha$ in IOU loss            & 10           \\
    The smooth parameter in IOU loss  & 1         \\
      $\alpha$ in Pixel Wise MSE loss & 20\\
       $\alpha$ in Generator loss & 100\\
        $\alpha$ in CBAM Generator loss & 1\\
         $\beta$ in CBAM Generator loss & 1\\
         $\alpha$ in Adaptive CBAM Generator loss & 3\\
    \bottomrule
  \end{tabular}
  \label{loss_function_parameters}
\end{table}

The parameters used in the RRT path planning algorithm are shown in \autoref{rrt_algorithm_parameters}:

\begin{table}[h]
 
  \caption{RRT Algorithm Parameters}
  \label{sample-table1}
  \centering
  \begin{tabular}{ll}
  \toprule
    Parameter     & Value           \\
    \midrule
    Step Length            & 0.2           \\
    Maximum Iterations             & 5000         \\
     Resolution of path & 1\\
    \bottomrule
  \end{tabular}
  \label{rrt_algorithm_parameters}
\end{table}

\section{Results}

Our GAN model outperforms the GAN model by \cite{zhang2021generative} using IOU, Dice, and FID as the image quality evaluation metrics. Moreover, our model has a smaller number of parameters and thus takes less time for both training and inference, which makes it a better choice for use as a heuristic for a sampling-based path planning algorithm, as shown in \autoref{quantitative_results}:

\begin{table}[h]
  \caption{Comparison of our results vs state-of-the-art networks using IOU, Dice, FID, and the number of parameters as the evaluation metrics.}
  \label{quantitative_results}
  \centering
  \begin{tabular}{lllll}
  \toprule
    Metrics     & IOU & Dice & FID  & Params       \\
    \midrule
    SAGAN \cite{zhang2021generative} & 88.45 & 93.88 & 66.47 & 1.24 Million    \\
    Our Network & \textbf{89.22} & \textbf{94.26} & \textbf{62.04}  & \textbf{0.88 Million}    \\
    \bottomrule
  \end{tabular}
\end{table}

The qualitative comparison of our approach versus the state-of-the-art network architecture in terms of the promising region generated is shown in \autoref{qualitative_results}:

\begin{figure}[h]
    \centering
    \includegraphics[width=6cm]{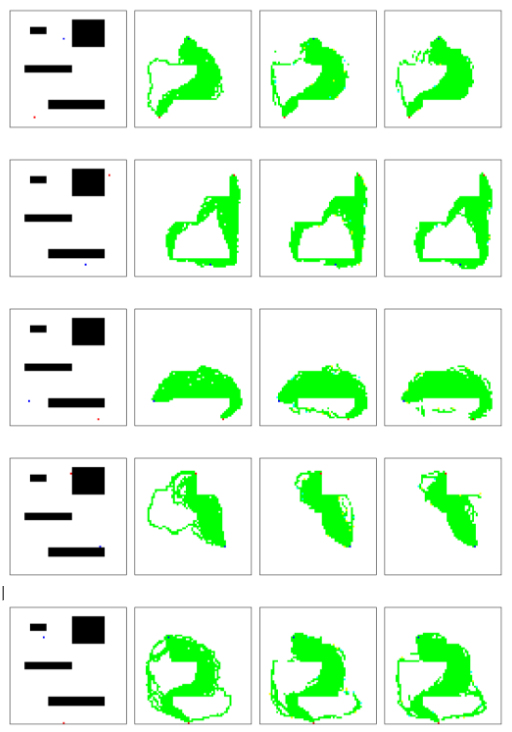}
    \caption{Illustration of the qualitative results. The first column shows the obstacle map with the white region as the free space and the black region as the obstacle space. The second column shows the promising region of the ground truth maps. The third column shows the promising region generated using the Self-Attention Generative Adversarial Network used in \cite{zhang2021generative}. The fourth column shows the promising region generated using our method. The red and the black nodes in all the maps denote the start and goal nodes, respectively.}
    \label{qualitative_results}
\end{figure}

Our model outperforms the previous state-of-the-art network architecture using both the time cost and the number of nodes as the path planning evaluation metric. The comparison of using various algorithms like the original RRT algorithm, SAGAN-RRT, and our CBAGAN-RRT network as the heuristic for the path planning algorithm using time cost and the number of nodes is shown in \autoref{quantitative_planning_results}: 

\begin{table}[h]

  \caption{Comparison of our method versus RRT and SAGAN-RRT on 3 different maps using time cost and the number of nodes as the path planning evaluation metrics.}
  \label{sample-table1}
  \centering
  \begin{tabular}{llll}
  \toprule
Map & Algorithm & Time Cost(s) & No. of Nodes  \\
    \midrule
 Map1 &RRT & 2.12 & 202\\
 Map1 &SAGAN \cite{zhang2021generative} & 1.37 & 157   \\
 Map1 &Our Network & \textbf{1.26} & \textbf{143}       \\
\midrule
  Map2 &RRT & 3.65 & 351\\
 Map2 &SAGAN \cite{zhang2021generative} & 2.40 & 246         \\
 Map2 &Our Network & \textbf{2.08} & \textbf{218}        \\
\midrule
  Map3 &RRT & 1.85 & 184\\
 Map3 &SAGAN \cite{zhang2021generative} & 1.28 & 129     \\
 Map3 &Our Network & \textbf{1.06} & \textbf{110}       \\
    \bottomrule
  \end{tabular}
  \label{quantitative_planning_results}
\end{table}

The qualitative comparison of our results using CBAGAN-RRT versus the original RRT in terms of the path generated is shown in \autoref{qualitative_planning_results}:

\begin{figure}[h]
    \centering
    \includegraphics[width=8cm]{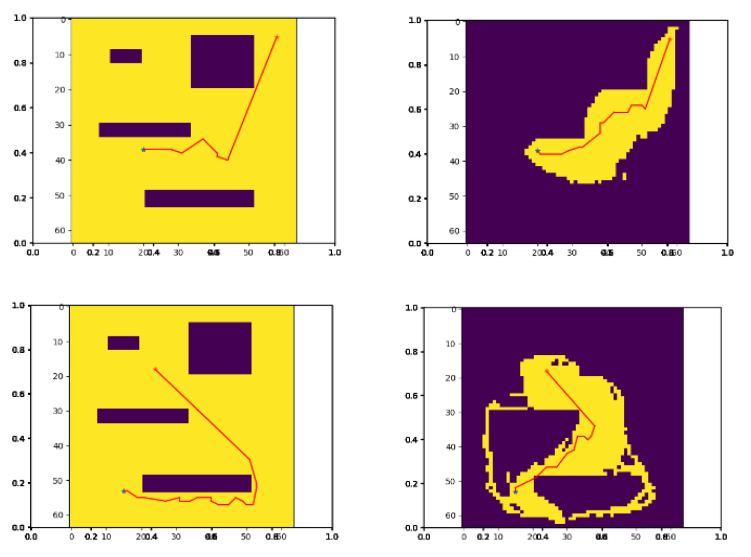}
    \caption{Illustration of our results using CBAGAN-RRT versus the original RRT in terms of the path generated. The left image depicts the path generated on the original obstacle map, where the obstacles are shown in the dark color, and the light color denotes the free space. The right image depicts the path generated using only the region of interest generated from our GAN model, where the lighter color denotes the area inside the region of interest, while the darker color denotes the region outside the region of interest. Only the region of interest is used for finding the feasible path, and because the area is much smaller than that in the other image, we can improve the speed of convergence of the algorithm. The start and the goal nodes are shown in color blue and red, respectively, while the red line denotes the path taken.}
    \label{qualitative_planning_results}
\end{figure}

\subsection{Ablation Study}

We conduct experiments using all the combinations of using spatial and channel attention modules and noticed that the best results using IOU, Dice, and FID as the image quality metrics and the time cost and number of nodes as the path planning metrics were achieved by using a combination of both spatial and channel attention modules in our network architecture as shown in \autoref{ablation_attention}:

\begin{table}[h]
  \caption{Comparison of our results with a combination of spatial and channel attention modules using IOU, Dice, and FID as the image quality evaluation metrics, and time cost and the number of nodes as the path planning metrics. Here, SA and CA denote Spatial Attention and Channel Attention, respectively.}
  \label{sample-table1}
  \centering
  \begin{tabular}{lllllll}
  \toprule
    SA & CA    & IOU & Dice & FID & Time Cost & No. of Nodes           \\
    \midrule
 $\checkmark$ & $\checkmark$ &  \textbf{89.22} & \textbf{94.26} & \textbf{62.04} &  \textbf{1.08} & \textbf{127}   \\
$\checkmark$ & $\times$ & 87.56 & 93.37 & 75.74 & 2.02 & 193\\    
$\times$  &  $\checkmark$ & 88.14 & 93.70 & 66.49 & 1.75 & 166 \\   
$\times$ & $\times$ & 87.25 & 93.19 & 82.53 & 2.56 & 235 \\  
    \bottomrule
    \label{ablation_attention}
  \end{tabular}
\end{table}

\section{Conclusions and Future Work}

In this paper, we proposed a novel image-based approach using a convolutional block attention-based Generative Adversarial Network network CBAGAN-RRT to design the heuristics and find better optimal solutions and improve the convergence, thus reducing the computational requirements for sampling-based path planning algorithms. We use the predicted probability distribution of paths generated from our model to guide the sampling process of the RRT algorithm. We demonstrate that our approach performs better than the state-of-the-art approach in terms of the quality of feasible paths generated from an image perspective using IOU, Dice, and FID as the evaluation metrics, as well as metrics like time cost and the number of nodes explored, from the path planning perspective. Our algorithm differs only in the sampling strategy from the RRT algorithm; hence, it can be easily integrated with other sampling-based path planning algorithms. A future idea could be to solve the problem by mapping it only as image segmentation and not image generation, that is, segmenting the feasible region using a state-of-the-art image segmentation algorithm and feeding that as heuristics for the path planning to improve the results. Another idea could be to integrate our GAN model with other sampling-based path-planning algorithms to study, test, and validate the effectiveness of our approach.

\nocite{*}
{\small
\bibliographystyle{splncs04}
\bibliography{egbib}

\begin{thebibliography}{10}
\providecommand{\url}[1]{\texttt{#1}}
\providecommand{\urlprefix}{URL }
\providecommand{\doi}[1]{https://doi.org/#1}

\bibitem{arslan2015machine}
Arslan, O., Tsiotras, P.: Machine learning guided exploration for sampling-based motion planning algorithms. In: 2015 IEEE/RSJ International Conference on Intelligent Robots and Systems (IROS). pp. 2646--2652. IEEE (2015)

\bibitem{burns2005sampling}
Burns, B., Brock, O.: Sampling-based motion planning using predictive models. In: Proceedings of the 2005 IEEE international conference on robotics and automation. pp. 3120--3125. IEEE (2005)

\bibitem{dang2022deep}
Dang, X., Chrpa, L., Edelkamp, S.: Deep rrt. In: Proceedings of the International Symposium on Combinatorial Search. vol.~15, pp. 333--335 (2022)

\bibitem{elbanhawi2014sampling}
Elbanhawi, M., Simic, M.: Sampling-based robot motion planning: A review. Ieee access  \textbf{2},  56--77 (2014)

\bibitem{fu2019dual}
Fu, J., Liu, J., Tian, H., Li, Y., Bao, Y., Fang, Z., Lu, H.: Dual attention network for scene segmentation. In: Proceedings of the IEEE/CVF conference on computer vision and pattern recognition. pp. 3146--3154 (2019)

\bibitem{gammell2014informed}
Gammell, J.D., Srinivasa, S.S., Barfoot, T.D.: Informed rrt: Optimal sampling-based path planning focused via direct sampling of an admissible ellipsoidal heuristic. In: 2014 IEEE/RSJ International Conference on Intelligent Robots and Systems. pp. 2997--3004. IEEE (2014)

\bibitem{gan2021research}
Gan, Y., Zhang, B., Ke, C., Zhu, X., He, W., Ihara, T.: Research on robot motion planning based on rrt algorithm with nonholonomic constraints. Neural Processing Letters  \textbf{53},  3011--3029 (2021)

\bibitem{goodfellow2016deep}
Goodfellow, I., Bengio, Y., Courville, A.: Deep learning. MIT press (2016)

\bibitem{goodfellow2014generative}
Goodfellow, I., Pouget-Abadie, J., Mirza, M., Xu, B., Warde-Farley, D., Ozair, S., Courville, A., Bengio, Y.: Generative adversarial nets. In: Advances in neural information processing systems. pp. 2672--2680 (2014)

\bibitem{hu2018squeeze}
Hu, J., Shen, L., Sun, G.: Squeeze-and-excitation networks. In: Proceedings of the IEEE conference on computer vision and pattern recognition. pp. 7132--7141 (2018)

\bibitem{isola2017image}
Isola, P., Zhu, J.Y., Zhou, T., Efros, A.A.: Image-to-image translation with conditional adversarial networks. In: Proceedings of the IEEE conference on computer vision and pattern recognition. pp. 1125--1134 (2017)

\bibitem{jeong2015rrt}
Jeong, I.B., Lee, S.J., Kim, J.H.: Rrt*-quick: A motion planning algorithm with faster convergence rate. In: Robot Intelligence Technology and Applications 3: Results from the 3rd International Conference on Robot Intelligence Technology and Applications. pp. 67--76. Springer (2015)

\bibitem{jeong2019quick}
Jeong, I.B., Lee, S.J., Kim, J.H.: Quick-rrt*: Triangular inequality-based implementation of rrt* with improved initial solution and convergence rate. Expert Systems with Applications  \textbf{123},  82--90 (2019)

\bibitem{karaman2011sampling}
Karaman, S., Frazzoli, E.: Sampling-based algorithms for optimal motion planning. The international journal of robotics research  \textbf{30}(7),  846--894 (2011)

\bibitem{karaman2011anytime}
Karaman, S., Walter, M.R., Perez, A., Frazzoli, E., Teller, S.: Anytime motion planning using the rrt. In: 2011 IEEE international conference on robotics and automation. pp. 1478--1483. IEEE (2011)

\bibitem{kuffner2000rrt}
Kuffner, J.J., LaValle, S.M.: Rrt-connect: An efficient approach to single-query path planning. In: Proceedings 2000 ICRA. Millennium Conference. IEEE International Conference on Robotics and Automation. Symposia Proceedings (Cat. No. 00CH37065). vol.~2, pp. 995--1001. IEEE (2000)

\bibitem{kuwata2008motion}
Kuwata, Y., Fiore, G.A., Teo, J., Frazzoli, E., How, J.P.: Motion planning for urban driving using rrt. In: 2008 IEEE/RSJ International Conference on Intelligent Robots and Systems. pp. 1681--1686. IEEE (2008)

\bibitem{lavalle2006planning}
LaValle, S.M.: Planning algorithms. Cambridge university press (2006)

\bibitem{lavalle2004relationship}
LaValle, S.M., Branicky, M.S., Lindemann, S.R.: On the relationship between classical grid search and probabilistic roadmaps. The International Journal of Robotics Research  \textbf{23}(7-8),  673--692 (2004)

\bibitem{lavalle2001rapidly}
LaValle, S.M., Kuffner, J.J.: Rapidly-exploring random trees: Progress and prospects: Steven m. lavalle, iowa state university, a james j. kuffner, jr., university of tokyo, tokyo, japan. Algorithmic and computational robotics pp. 303--307 (2001)

\bibitem{lavalle2001randomized}
LaValle, S.M., Kuffner~Jr, J.J.: Randomized kinodynamic planning. The international journal of robotics research  \textbf{20}(5),  378--400 (2001)

\bibitem{lavalle1998rapidly}
LaValle, S.M., et~al.: Rapidly-exploring random trees: A new tool for path planning  (1998)

\bibitem{li2019spatial}
Li, X., Hu, X., Yang, J.: Spatial group-wise enhance: Improving semantic feature learning in convolutional networks. arXiv preprint arXiv:1905.09646  (2019)

\bibitem{li2019selective}
Li, X., Wang, W., Hu, X., Yang, J.: Selective kernel networks. In: Proceedings of the IEEE/CVF conference on computer vision and pattern recognition. pp. 510--519 (2019)

\bibitem{li2018neural}
Li, Y., Cui, R., Li, Z., Xu, D.: Neural network approximation based near-optimal motion planning with kinodynamic constraints using rrt. IEEE Transactions on Industrial Electronics  \textbf{65}(11),  8718--8729 (2018)

\bibitem{li2022contextual}
Li, Y., Yao, T., Pan, Y., Mei, T.: Contextual transformer networks for visual recognition. IEEE Transactions on Pattern Analysis and Machine Intelligence  (2022)

\bibitem{ma2022enhance}
Ma, H., Li, C., Liu, J., Wang, J., Meng, M.Q.H.: Enhance connectivity of promising regions for sampling-based path planning. IEEE Transactions on Automation Science and Engineering  (2022)

\bibitem{ma2015efficient}
Ma, L., Xue, J., Kawabata, K., Zhu, J., Ma, C., Zheng, N.: Efficient sampling-based motion planning for on-road autonomous driving. IEEE Transactions on Intelligent Transportation Systems  \textbf{16}(4),  1961--1976 (2015)

\bibitem{ma2021conditional}
Ma, N., Wang, J., Liu, J., Meng, M.Q.H.: Conditional generative adversarial networks for optimal path planning. IEEE Transactions on Cognitive and Developmental Systems  \textbf{14}(2),  662--671 (2021)

\bibitem{mcmahon2022survey}
McMahon, T., Sivaramakrishnan, A., Granados, E., Bekris, K.E., et~al.: A survey on the integration of machine learning with sampling-based motion planning. Foundations and Trends{\textregistered} in Robotics  \textbf{9}(4),  266--327 (2022)

\bibitem{misra2021rotate}
Misra, D., Nalamada, T., Arasanipalai, A.U., Hou, Q.: Rotate to attend: Convolutional triplet attention module. In: Proceedings of the IEEE/CVF Winter Conference on Applications of Computer Vision. pp. 3139--3148 (2021)

\bibitem{naderi2015rt}
Naderi, K., Rajam{\"a}ki, J., H{\"a}m{\"a}l{\"a}inen, P.: Rt-rrt* a real-time path planning algorithm based on rrt. In: Proceedings of the 8th ACM SIGGRAPH Conference on Motion in Games. pp. 113--118 (2015)

\bibitem{nasir2013rrt}
Nasir, J., Islam, F., Malik, U., Ayaz, Y., Hasan, O., Khan, M., Muhammad, M.S.: Rrt*-smart: A rapid convergence implementation of rrt. International Journal of Advanced Robotic Systems  \textbf{10}(7), ~299 (2013)

\bibitem{noreen2016comparison}
Noreen, I., Khan, A., Habib, Z.: A comparison of rrt, rrt* and rrt*-smart path planning algorithms. International Journal of Computer Science and Network Security (IJCSNS)  \textbf{16}(10), ~20 (2016)

\bibitem{noreen2016optimal}
Noreen, I., Khan, A., Habib, Z.: Optimal path planning using rrt* based approaches: a survey and future directions. International Journal of Advanced Computer Science and Applications  \textbf{7}(11) (2016)

\bibitem{perez2012lqr}
Perez, A., Platt, R., Konidaris, G., Kaelbling, L., Lozano-Perez, T.: Lqr-rrt*: Optimal sampling-based motion planning with automatically derived extension heuristics. In: 2012 IEEE International Conference on Robotics and Automation. pp. 2537--2542. IEEE (2012)

\bibitem{qi2020mod}
Qi, J., Yang, H., Sun, H.: Mod-rrt*: A sampling-based algorithm for robot path planning in dynamic environment. IEEE Transactions on Industrial Electronics  \textbf{68}(8),  7244--7251 (2020)

\bibitem{salzman2016asymptotically}
Salzman, O., Halperin, D.: Asymptotically near-optimal rrt for fast, high-quality motion planning. IEEE Transactions on Robotics  \textbf{32}(3),  473--483 (2016)

\bibitem{strub2022adaptively}
Strub, M.P., Gammell, J.D.: Adaptively informed trees (ait*) and effort informed trees (eit*): Asymmetric bidirectional sampling-based path planning. The International Journal of Robotics Research  \textbf{41}(4),  390--417 (2022)

\bibitem{tahir2018potentially}
Tahir, Z., Qureshi, A.H., Ayaz, Y., Nawaz, R.: Potentially guided bidirectionalized rrt* for fast optimal path planning in cluttered environments. Robotics and Autonomous Systems  \textbf{108},  13--27 (2018)

\bibitem{tsianos2007sampling}
Tsianos, K.I., Sucan, I.A., Kavraki, L.E.: Sampling-based robot motion planning: Towards realistic applications. Computer Science Review  \textbf{1}(1),  2--11 (2007)

\bibitem{vaswani2017attention}
Vaswani, A., Shazeer, N., Parmar, N., Uszkoreit, J., Jones, L., Gomez, A.N., Kaiser, {\L}., Polosukhin, I.: Attention is all you need. Advances in neural information processing systems  \textbf{30} (2017)

\bibitem{wang2021deep}
Wang, J., Jia, X., Zhang, T., Ma, N., Meng, M.Q.H.: Deep neural network enhanced sampling-based path planning in 3d space. IEEE Transactions on Automation Science and Engineering  \textbf{19}(4),  3434--3443 (2021)

\bibitem{wang2022gmr}
Wang, J., Li, T., Li, B., Meng, M.Q.H.: Gmr-rrt*: Sampling-based path planning using gaussian mixture regression. IEEE Transactions on Intelligent Vehicles  \textbf{7}(3),  690--700 (2022)

\bibitem{wang2020eca}
Wang, Q., Wu, B., Zhu, P., Li, P., Zuo, W., Hu, Q.: Eca-net: Efficient channel attention for deep convolutional neural networks. In: Proceedings of the IEEE/CVF conference on computer vision and pattern recognition. pp. 11534--11542 (2020)

\bibitem{webb2013kinodynamic}
Webb, D.J., Van Den~Berg, J.: Kinodynamic rrt*: Asymptotically optimal motion planning for robots with linear dynamics. In: 2013 IEEE international conference on robotics and automation. pp. 5054--5061. IEEE (2013)

\bibitem{wei2018method}
Wei, K., Ren, B.: A method on dynamic path planning for robotic manipulator autonomous obstacle avoidance based on an improved rrt algorithm. Sensors  \textbf{18}(2), ~571 (2018)

\bibitem{woo2018cbam}
Woo, S., Park, J., Lee, J.Y., Kweon, I.S.: Cbam: Convolutional block attention module. In: Proceedings of the European conference on computer vision (ECCV). pp. 3--19 (2018)

\bibitem{wu2021fast}
Wu, Z., Meng, Z., Zhao, W., Wu, Z.: Fast-rrt: A rrt-based optimal path finding method. Applied Sciences  \textbf{11}(24),  11777 (2021)

\bibitem{xinyu2019bidirectional}
Xinyu, W., Xiaojuan, L., Yong, G., Jiadong, S., Rui, W.: Bidirectional potential guided rrt* for motion planning. IEEE Access  \textbf{7},  95046--95057 (2019)

\bibitem{xiong2020rapidly}
Xiong, C., Zhou, H., Lu, D., Zeng, Z., Lian, L., Yu, C.: Rapidly-exploring adaptive sampling tree*: a sample-based path-planning algorithm for unmanned marine vehicles information gathering in variable ocean environments. Sensors  \textbf{20}(9), ~2515 (2020)

\bibitem{yu2021reducing}
Yu, C., Gao, S.: Reducing collision checking for sampling-based motion planning using graph neural networks. Advances in Neural Information Processing Systems  \textbf{34},  4274--4289 (2021)

\bibitem{zhang2019self}
Zhang, H., Goodfellow, I., Metaxas, D., Odena, A.: Self-attention generative adversarial networks. In: International conference on machine learning. pp. 7354--7363. PMLR (2019)

\bibitem{zhang2021sa}
Zhang, Q.L., Yang, Y.B.: Sa-net: Shuffle attention for deep convolutional neural networks. In: ICASSP 2021-2021 IEEE International Conference on Acoustics, Speech and Signal Processing (ICASSP). pp. 2235--2239. IEEE (2021)

\bibitem{zhang2021generative}
Zhang, T., Wang, J., Meng, M.Q.H.: Generative adversarial network based heuristics for sampling-based path planning. IEEE/CAA Journal of Automatica Sinica  \textbf{9}(1),  64--74 (2021)

\end{thebibliography}
}

\end{document}